\documentclass[acmtog,nonacm=true]{acmart}

\def\BibTeX{{\rm B\kern-.05em{\sc i\kern-.025em b}\kern-.08emT\kern-.1667em\lower.7ex\hbox{E}\kern-.125emX}}
\setcopyright{none}
\usepackage{color,xcolor}
\usepackage{epsfig}
\usepackage{graphicx}

\usepackage{tikz}
\usetikzlibrary{arrows}

\usepackage{tabularx}
\usepackage{adjustbox}
\usepackage{array}
\usepackage{booktabs}
\usepackage{colortbl}
\usepackage{float,wrapfig}
\usepackage{hhline}
\usepackage{multirow}
\usepackage{subcaption} %

\usepackage{amsmath,amsfonts,amsthm,amssymb}
\usepackage{bm}
\usepackage{nicefrac}
\usepackage{microtype}
\usepackage{dsfont}
\usepackage{changepage}
\usepackage{extramarks}
\usepackage{fancyhdr}
\usepackage{setspace}
\usepackage{soul}
\usepackage{xspace}

\usepackage{url}

\usepackage{algorithm, algorithmic}
\usepackage{enumitem}
\usepackage{todonotes} %

\usepackage{amsmath,amsfonts,bm}

\def\rx{{\textnormal{x}}}

\def\rz{{\textnormal{z}}}

\DeclareMathAlphabet{\mathsfit}{\encodingdefault}{\sfdefault}{m}{sl}
\SetMathAlphabet{\mathsfit}{bold}{\encodingdefault}{\sfdefault}{bx}{n}

\usepackage{amsmath,amsfonts,bm}

\newcommand{\edit}{\texttt{edit}}

\newcommand{\true}{\mathrm{true}}

\newcommand{\Enet}{E_\mathrm{net}}

\newcommand{\method}{image-specific generative model\xspace}
\newcommand{\methodadj}{image-specific\xspace}

\newcommand{\guidance}{reference image\xspace}

\newcommand{\mask}{\texttt{mask}}

\newcommand{\reffig}[1]{Figure~\ref{fig:#1}}
\newcommand{\refsec}[1]{Section~\ref{sec:#1}}

\newcommand{\refeq}[1]{Eqn.~\ref{eq:#1}}

\newcommand{\lblfig}[1]{\label{fig:#1}}
\newcommand{\lblsec}[1]{\label{sec:#1}}
\newcommand{\lbleq}[1]{\label{eq:#1}}

\newcommand{\ignorethis}[1]{}

\newcommand{\myparagraph}[1]{\vspace{-0pt}\paragraph{#1}}

\def\eqref#1{equation~\ref{#1}}

\def\1{\bm{1}}

\def\rx{{\textnormal{x}}}

\def\rz{{\textnormal{z}}}

\DeclareMathAlphabet{\mathsfit}{\encodingdefault}{\sfdefault}{m}{sl}
\SetMathAlphabet{\mathsfit}{bold}{\encodingdefault}{\sfdefault}{bx}{n}

\newcommand{\Ls}{\mathcal{L}}
\newcommand{\R}{\mathbb{R}}

\newcolumntype{L}[1]{>{\raggedright\let\newline\\\arraybackslash\hspace{0pt}}m{#1}}
\newcolumntype{C}[1]{>{\centering\let\newline\\\arraybackslash\hspace{0pt}}m{#1}}
\newcolumntype{R}[1]{>{\raggedleft\let\newline\\\arraybackslash\hspace{0pt}}m{#1}}

\newcommand{\ignore}[1]{}

\makeatletter
\DeclareRobustCommand\onedot{\futurelet\@let@token\@onedot}
\def\@onedot{\ifx\@let@token.\else.\null\fi\xspace}

\makeatother

\definecolor{MyDarkBlue}{rgb}{0,0.08,1}
\definecolor{MyDarkGreen}{rgb}{0.02,0.6,0.02}
\definecolor{MyDarkRed}{rgb}{0.8,0.02,0.02}
\definecolor{MyDarkOrange}{rgb}{0.40,0.2,0.02}
\definecolor{MyPurple}{RGB}{111,0,255}
\definecolor{MyRed}{rgb}{1.0,0.0,0.0}
\definecolor{MyGold}{rgb}{0.75,0.6,0.12}
\definecolor{MyDarkgray}{rgb}{0.66, 0.66, 0.66}

\setcopyright{rightsretained} 
\acmJournal{TOG}
\acmYear{2019}
\acmVolume{38}
\acmNumber{4}
\acmArticle{59}
\acmMonth{7}
\acmDOI{10.1145/3306346.3323023}

\begin{document}

\title{Semantic Photo Manipulation with a Generative Image Prior}

\author{David Bau}
\orcid{0000-0003-1744-6765}
\affiliation{%
  \institution{MIT CSAIL and MIT-IBM Watson AI Lab}
}
\email{davidbau@csail.mit.edu}

\author{Hendrik Strobelt}
\affiliation{%
  \institution{IBM Research and MIT-IBM Watson AI Lab}
}
\email{hendrik.strobelt@ibm.com}

\author{William Peebles}
\affiliation{%
  \institution{MIT CSAIL}
}
\email{wisp@csail.mit.edu}

\author{Jonas Wulff}
\affiliation{%
  \institution{MIT CSAIL}
}
\email{jwulff@csail.mit.edu}

\author{Bolei Zhou}
\affiliation{%
  \institution{The Chinese University of Hong Kong}
}
\orcid{0000-0003-4030-0684}
\email{bzhou@ie.cuhk.edu.hk}

\author{Jun-Yan Zhu}
\affiliation{%
  \institution{MIT CSAIL}
}
\email{junyanz@csail.mit.edu}

\author{Antonio Torralba}
\affiliation{%
  \institution{MIT CSAIL and MIT-IBM Watson AI Lab}
}
\email{torralba@csail.mit.edu}

\begin{abstract}
Despite the recent success of GANs in synthesizing images conditioned on inputs such as a user sketch, text, or semantic labels, manipulating the high-level attributes of an existing natural photograph with GANs is challenging for two reasons. First, it is hard for GANs to precisely reproduce an input image. Second, after manipulation, the newly synthesized pixels often do not fit the original image.  In this paper, we address these issues by adapting the image prior learned by GANs to image statistics of an individual image. Our method can accurately reconstruct the input image and synthesize new content, consistent with the appearance of the input image. We demonstrate our interactive system on several semantic image editing tasks, including synthesizing new objects consistent with background, removing unwanted objects, and changing the appearance of an object. Quantitative and qualitative comparisons against several existing methods demonstrate the effectiveness of our method.
\end{abstract}



\begin{teaserfigure}
\includegraphics[width=\textwidth]{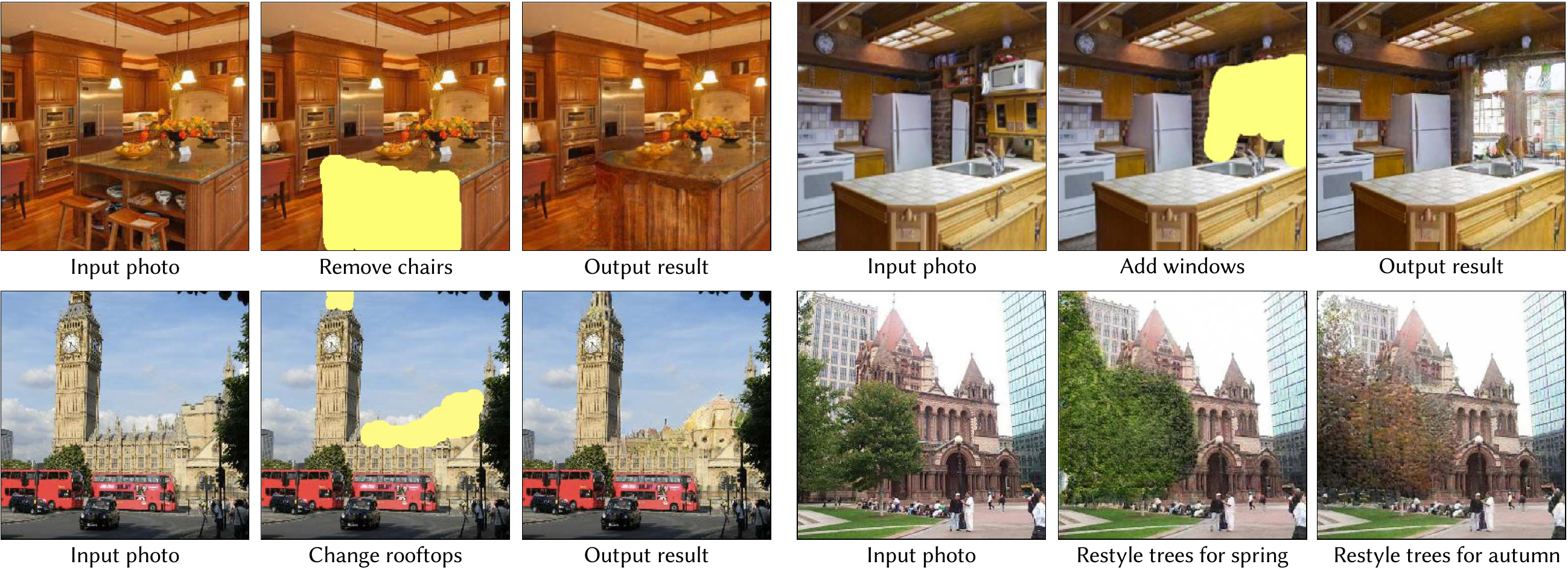}
\caption{Our proposed method enables several new interactive photo manipulations in which a user edits a photo with high-level concepts rather than pixel colors. Our deep generative model can synthesize new content that follows both the user's intention and the natural image statistics. {\emph Top:} Given simple user strokes, our method can automatically synthesize and manipulate different objects while adjusting the surrounding context to match. {\emph Bottom:} Our users can edit the visual appearance of objects directly, such as changing the appearance of rooftops or trees. Photos from the LSUN dataset~\citep{yu2015lsun}.}
\label{fig:teaser}
\end{teaserfigure}

\maketitle
\section{Introduction}
\lblsec{intro}

The whirlwind of progress in deep learning has produced a steady stream of promising generative models~\cite{goodfellow2014generative, karras2018progressive} that render natural scenes increasingly indistinguishable from reality and provide an intuitive way to generate realistic imagery given high-level user inputs~\cite{wang2018pix2pixHD,bau2019gandissect}.

\begin{figure*}
\includegraphics[width=0.8\textwidth]{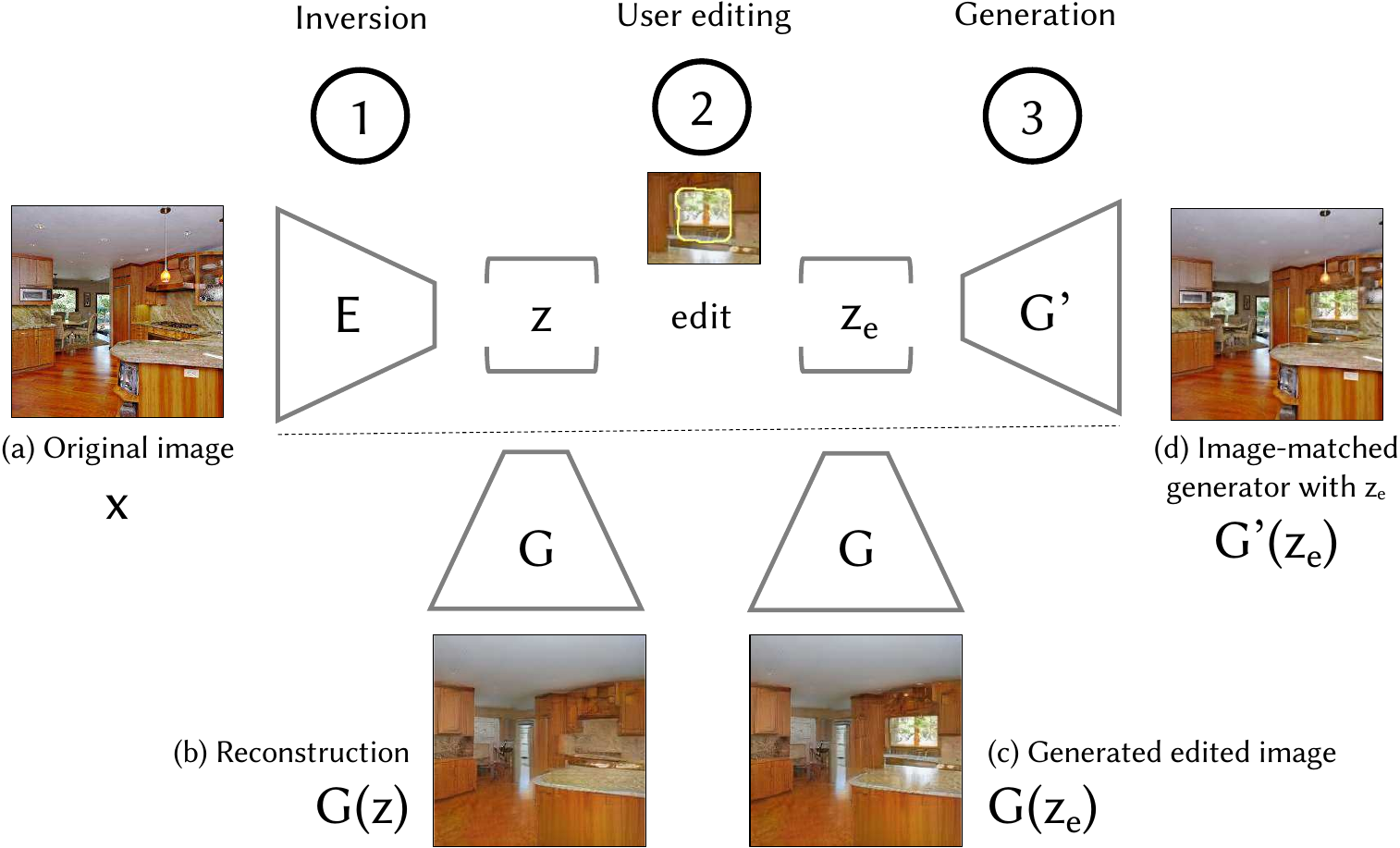}
\caption{\emph{Overview.} To perform a semantic edit on an image $\rx$, we take three steps. (1) We first compute a latent vector $\rz=E(\rx)$ representing $\rx$. (2) We then apply a semantic vector space operation $\rz_e = \edit(\rz)$ in the latent space; this could add, remove, or alter a semantic concept in the image. (3) Finally, we regenerate the image from the modified $\rz_e$.  Unfortunately, as can be seen in (b), usually the input image $\rx$ cannot be precisely generated by the generator $G$, so (c) using the generator $G$ to create the edited image $G(\rx_e)$ will result in the loss of many attributes and details of the original image (a). Therefore to generate the image we propose a new last step: (d) We learn an \methodadj generator $G'$ which can produce $\rx'_e = G'(z_e)$ that is faithful to the original image $\rx$ in the unedited regions.   Photo from the LSUN dataset~\citep{yu2015lsun}.}

\Description{Ideally the variable $\rx_e$ will end up similar to $\rx$, but the generator $G$ will typically generate images that are not similar to $\rx$.  Our method tunes $G'$ so that $\rx'_e = G'(\rz_e)$ is faithful to $\rx$.}
\lblfig{framework}
\end{figure*}

We seem to be on the verge of using these generative models for \textit{semantic manipulation of natural photographs}, which will combine the simplicity of pixel-level editing with flexible semantic manipulations of a scene.  The role of deep generative models will be to provide latent semantic representations in which concepts can be directly manipulated and then to preserve image realism when semantic changes are made. This editing scheme will allow users to manipulate a photograph not with physical colors, but with abstract concepts such as object types and visual attributes. For example, a user can add a lit lamp into a bedroom (\reffig{teaser} bottom left) or change the color of a tree's leaves (\reffig{teaser} bottom right) with a few scribbles.

Despite the promise of this approach, two  technical challenges have prevented these generative models from being applied to natural photograph manipulation. First, it is extremely difficult to find a latent code $z$  to reproduce a given photograph $x$ with a deep generative model $G$: $\rx \approx G(\rz)$. As shown in  ~\reffig{framework}b, the reconstructed image $G(\rz)$ roughly captures the visual content of the input image $x$, but visual details are obviously different from the original photo. Second, after manipulation, the newly synthesized pixels from generative models are often incompatible with the existing content from the real image, which makes stitching the new content into the context of the original image challenging~(\reffig{framework}c).

In this paper, we address the above two issues using an image-specific adaptation method.  Our key idea is to learn an \textit{\method} $G' \approx G$, that produces a near-exact solution for our input image $\rx$, so that $\rx \approx G'(\rz)$ outside the edited region of the image. We construct our image-specific $G'$ to share the same semantic representations as of the original $G$.   
Importantly, our \methodadj $G'$ produces new visual content, consistent with the original photo while reflecting semantic manipulations (\reffig{framework}d).

We build \texttt{GANPaint editor}, an interactive interface that supports a wide range of editing tasks, from inserting new objects into a natural photo (\reffig{teaser} top)  to changing the attributes of existing objects (\reffig{teaser} bottom).
We show that our general-purpose editing method outperforms compositing-based methods as measured in human perception studies. Finally, we perform an ablation study demonstrating the importance of our image-specific adaptation method compared to previous reconstruction methods. 
Our code, models, and data are available at our website \url{ganpaint.csail.mit.edu}.

\section{Related Work}
\lblsec{related}
\myparagraph{Generative Adversarial Networks } (GANs)~\cite{goodfellow2014generative} learn to automatically synthesize realistic image samples~\cite{miyato2018spectral,karras2018progressive}. GANs have enabled several user-guided image synthesis tasks such as generating images from user sketches~\cite{isola2017image,sangkloy2016scribbler}, creating face animation~\cite{nagano2018pagan,geng2018warp}, synthesizing photos from language description~\cite{han2017stackgan}, inpainting~\cite{yeh2017semantic}, and interactively manipulating objects in a generated scene~\cite{bau2019gandissect,park2019SPADE}. While most of the prior work focuses on generating a new image \textit{from scratch} given user controls, little work has used GANs for interactively manipulating an existing natural photograph. The key challenge is the mismatch between the GAN-generated content and existing content in the image.  Several lines of work~\cite{zhu2016generative,perarnau2016invertible,brock2017neural} propose to manipulate a photo using GANs but only work with a single object (e.g., handbag) at low resolutions (64x64) and often involve post-processing steps. In this work, our method can directly generate a final result and allow semantic manipulations of an entire natural scene. 

\myparagraph{Interactive Photo Manipulation.} Manipulating a natural photograph is a central problem in computer graphics and computational photography, where a piece of software manipulates an image to achieve a user-specified goal while keeping the result photorealistic.  Example applications include color adjustment~\cite{reinhard2001color,levin2004colorization,an2008appprop,xue2012understanding}, tone mapping~\cite{durand2002fast}, image warping~\cite{avidan2007seam}, image blending~\cite{perez2003poisson,tao2010error}, or image reshuffling~\cite{barnes2009patchmatch}, to name just a few. These editing tools can modify the low-level features of an image with simple interactions (e.g. scribbles) and work well without external knowledge. On the contrary, to edit the high-level semantics of an image such as the presence and appearance of objects, prior work often demands manual annotations of the object geometry~\cite{kholgade20143d} and scene layout~\cite{karsch2011rendering},  choice of an appropriate object~\cite{perez2003poisson,lalonde2007photo}, RGBD data~\cite{zhang2016emptying}, or hierarchical segmentations~\cite{hong2018learning}. Different from those systems, our method allows complex high-level semantic manipulations of natural photos with a simple brush-based user interface. We use natural image statistics learned by a deep generative model to connect simple user interaction to the complex visual world.

\myparagraph{Deep Image Manipulation.}
Deep learning has achieved compelling results in many image editing tasks. Recent examples include image inpainting~\cite{pathak2016context,iizuka2017globally,yu2018inpainting}, image colorization~\cite{zhang2016colorful,iizuka2016let}, photo stylization~\cite{gatys2015neural,zhu2017unpaired,li2018closed} and photo enhancement~\cite{gharbi2017deep,kim2018deep}. Learning-based systems allow real-time computation and require no hand-crafted heuristics. Recent work further integrates user interaction into end-to-end learning systems, enabling interactive applications such as user-guided colorization~\cite{zhang2017real} and sketch-based face editing~\cite{Faceshop}.
  These methods can achieve high-quality results, but the editing task is fixed at training time and requires specific training data. In this work, we propose an alternative, task-agnostic approach. We learn the natural image statistics using a generative model and allow many different editing applications in the same framework. This enables new visual effects where training data is not available, such as adding objects and changing the appearance of objects.%

\begin{figure}
\includegraphics[width=.8\columnwidth]{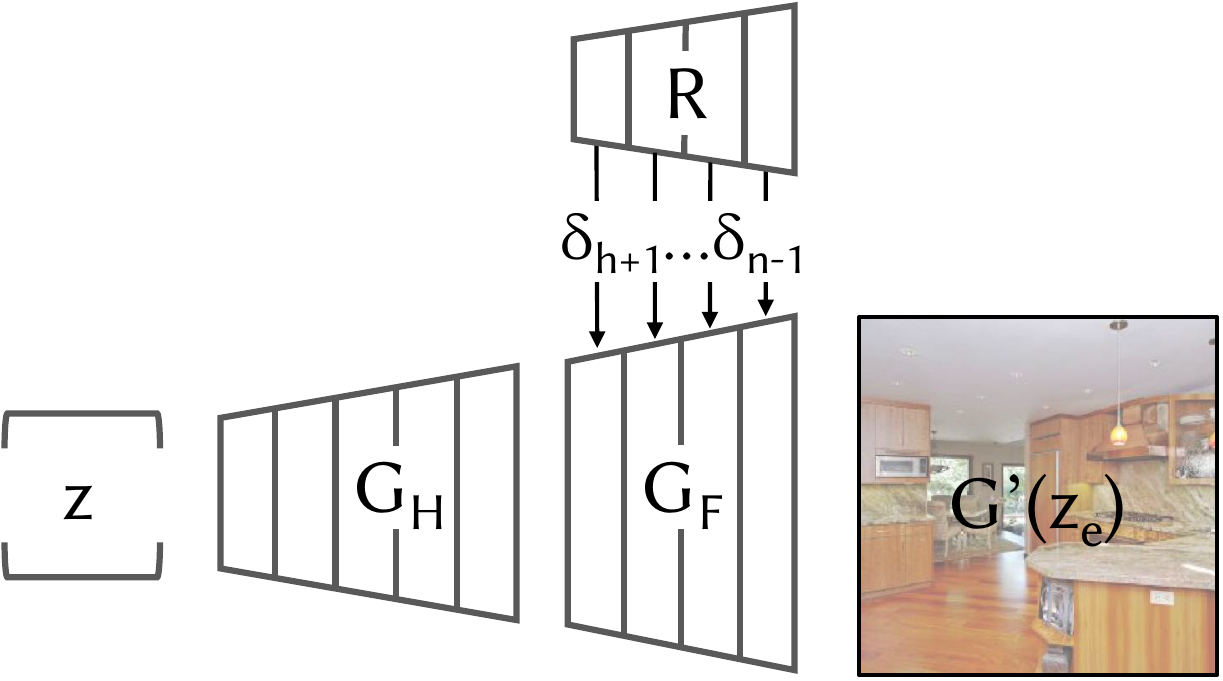}
\caption{Image-specific model adaptation through training a small network $R$ that produces perturbations $\delta_i$ that influence the later, fine-grained layers $G_F$ of the GAN. We add a regularization term to encourage the perturbations to be small.}
\lblfig{formula-adaptation}
\end{figure}

\section{Method}
\lblsec{method}
\myparagraph{Overview:} We propose a general-purpose semantic photo manipulation method that integrates the natural image prior captured by a GAN generator. \reffig{framework} shows our image editing pipeline. Given a natural photograph as input, we first re-render the image using an image generator. More concretely, to precisely reconstruct the input image, our method not only optimizes the latent representation but also adapts the generator. The user then manipulates the photo using our interactive interface, such as by adding or removing specific objects or changing their appearances. Our method updates the latent representation according to each edit and renders the final result given the modified representation. Our results look both realistic and visually similar to the input natural photograph.

Below, we first review recent GAN-based methods for generating visual content given semantic user inputs in \refsec{method-synthesis}. In theory, we could apply these methods to edit an existing photo as long as we can reproduce an image using a learned GAN generator. In \refsec{method-reconstruct}, we show that a rough reconstruction is possible, but a precise reconstruction is challenging and has eluded prior work.  In \refsec{method-adapt}, we present our new \methodadj adaptation method to bridge the gap between the original photo and generated content. Finally, we describe several image manipulation operations powered by our algorithm in \refsec{method-edit}.

\subsection{Controllable Image Synthesis with GANs}
\lblsec{method-synthesis}
Deep generative models~\cite{goodfellow2014generative,kingma2014auto} are proficient at learning meaningful latent representation spaces. An encoder-based generative model is a function $G: \rz \rightarrow \rx$ that generates an image $\rx \in \mathds{R}^{H\times W\times 3}$ from a latent representation $\rz \in \mathds{R}^{h\times w\times |z|}$. This representation can be a low-dimensional vector from a Gaussian distribution (i.e., $1\!\times 1\!\times |z|$) or an intermediate feature representation in the generator. In this work, we use the intermediate representation for flexible spatial control. 

Using a generative model $G$ for photo editing is powerful for two reasons. First, arithmetic vector operations in the latent representation space often result in interesting semantic manipulations.  For example, given a generated image from the model $G(\rz)$, latent vector operations $\rz_e = \edit(\rz)$ can adjust the species of an animal~\cite{miyato2018spectral}, the orientation of an object~\cite{chen2016infogan}, or the appearance~\cite{zhu2016generative} or presence~\cite{bau2019gandissect} of objects in a scene. Second, the edited image $G\left( \edit(\rz) \right)$ will still lie on the natural image manifold as $G$ is trained to produce a natural image given any latent code.

Our method can be applied to various generative models and different latent representation editing methods. In this work, we focus on manipulating semantic object-centric representations that can add, remove, and alter objects such as trees and doors within a natural scene. For that, we build our work on a state-of-the-art model, progressive GANs~\cite{karras2018progressive}, and a recent latent representation editing method~\cite{bau2019gandissect}. \reffig{teaser} shows several editing examples on real input images.  The details of these latent space edits are discussed in \refsec{method-edit}.%

\begin{figure}
\includegraphics[width=\columnwidth]{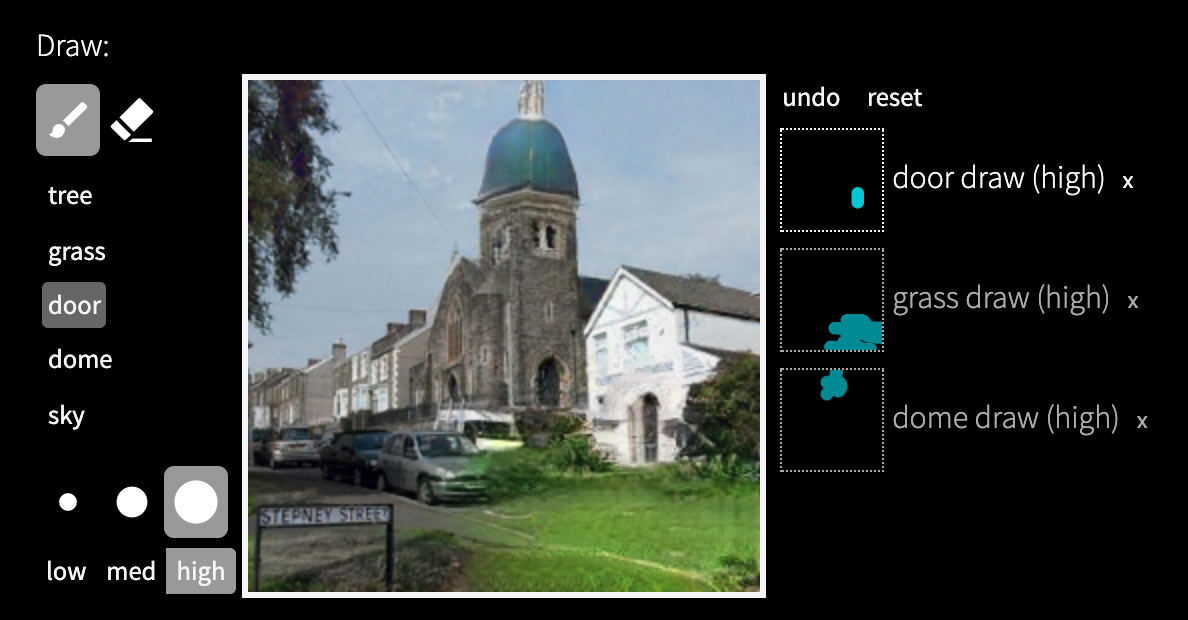}
\caption{GANBrush user interface. A new image is first uploaded and inverted.  The toolbar (left of image) then allows users to select the mode of operation (draw or erase), select the semantic feature, and select brush-size and feature strength (low, med, high). The history panel (right of image) shows a stack of modifications in chronological order. The corresponding edits are highlighted in the image when users hover over previous edits. Each edit can also be removed from the list.  Photo being edited is of St Luke the Evangelist, Cwmbwrla, Swansea courtesy Jaggery, via \href{https://www.geograph.org.uk/photo/3670928}{Geograph} (cc-by-sa/2.0).}
\lblfig{ui_screenshot}
\end{figure}

\subsection{Reproducing a Natural Image with a Generator}
\lblsec{method-reconstruct}
Applying a semantic edit as described above on a natural image requires reconstructing the input image $\rx$ in the latent space (i.e.~finding a $\rz$ so that $\rx \approx G(\rz)$), applying the semantic manipulation $\rz_e = \edit(\rz)$ in that space, and then using the generator $\rx_e=G(\rz_e)$ to render the modified image (\reffig{framework}). Formally, we seek for a latent code $\rz$ that minimizes the reconstruction loss $\mathcal{L}_{r}(\rx, G(\rz))$ between the input image $\rx$ and generated image $G(\rz)$: 
\begin{align}
\mathcal{L}_{r}(\rx, G(\rz)) = 
\left\Vert \rx  - G(\rz) \right\Vert_1 +  \lambda_{\text{VGG}} \sum_{i=1}^N\frac{1}{M_i} \Vert F^{(i)}(\rx)-F^{(i)}(G(\rz)) \Vert_1,
\lbleq{opt}
\end{align}
where we use both color pixel loss and perceptual loss, similar to prior image reconstruction work~\cite{dosovitskiy2016generating,zhu2016generative}. Here $\lambda_{\text{VGG}}=10$ and $F^{(i)}$ is the $i$-th layer with $M_i$ features in the VGG network~\cite{simonyan2014very}. To speed up the reconstruction, we follow the prior work~\cite{zhu2016generative} and train an encoder $E\colon  \rx \rightarrow \rz$ that predicts the latent code directly from the image. We use a similar training loss $\arg \min_E \mathbb{E}_{\rx\sim p_{data}(\rx)} \mathcal{L}_{r} (\rx, G(E(\rx)))$. During test time, we use $E(\rx)$ as the initialization for the optimization of \refeq{opt}.

Unfortunately, this method fails to produce visually appealing edits when $G$ is not able to generate images resembling $\rx$. Finding a latent code $\rz$  that can reproduce an \textit{arbitrary image} $x$ is hard because, for many images, the range of the generator $G$ does not include any image sufficiently similar to $x$ in appearance. As shown in \reffig{framework}, existing reconstruction methods~\cite{zhu2016generative,dosovitskiy2016generating} can only roughly re-generate the color and shape of objects in the scene and fail to reproduce the visual details of input images faithfully. Given an inaccurate reconstruction, subsequent edits introduce further artifacts. Besides, the unedited regions of the final result may look different from the input photo due to the reconstruction error.

\subsection{Image-Specific Adaptation}
\lblsec{method-adapt}

The image reconstruction problem is hard as it would require the same generator $G$ to be able to reproduce every single detail of every possible input image. And to be useful for incremental editing, new results $G(\edit(\rz))$ need to be compatible with the input image $\rx$ as well.  To address these issues, we propose to use an \methodadj generator $G'$ that can adapt itself to a particular image. %
First, this \method $G'$ can produce a near-exact match for our input image $\rx$. %
Second, our image-specific $G'$ should be close to $G$ so that they share an underlying semantic representation.
Learned with the above two objectives,  $G'$ can preserve the visual details of the original photo during semantic manipulations.

More precisely, to perform a successful edit, $G'$ does not strictly need to have $\rx$ itself in its range; rather, we find $G'$ to exactly reproduce only the unedited regions of the input image. Given a user stroke binary mask, $\mask_e$: 
 \begin{align}
 \mask_e = \begin{cases}
 1 & \text{where the stroke is present} \\
 0 & \text{outside the stroke}
 \end{cases}
 \end{align}
When adapting $G'$, this constraint can be approximated by minimizing a simple difference between the input image $\rx$ and those generated by $G'(\rz_e)$, summed over the image regions outide of the strokes.
\begin{align}
\Ls_{\text{match}} \equiv 
|| (G'(\rz_e) - x )\odot (1 - \mask_e) ||_1,
\end{align}
where we set $\rz_e = \edit(\rz)$, and $\odot$ is the elementwise Hadamard product. %
The operation $\edit(\rz)$ expresses the user's intent to apply a particular semantic manipulation on the deep latent structure of $G$; it assumes that we can find a $G'$ with a similar latent structure as $G$. Otherwise, the editing operation $\edit(\rz)$ may not work well for newly constructed $G'$.

\myparagraph{Preserving Semantic Representation.}
To ensure that the \methodadj generator $G'$ has a similar latent space structure as the original generator $G$, we construct $G'$ by preserving all the early layers of $G$ precisely and applying perturbations only at the layers of the network that determine the fine-grained details.

This is done by exploiting the internal structure of modern image  generator networks: a generator $G$ has a layered structure consisting of a series of convolutions at increasing resolutions, where the final layer $g_n$ is close to pixels and captures the fine-grained details, while the first layer $g_1$ is closest to the latent representation $\rz$ and captures high-level information:
\begin{align}
G(z) = g_{n}(g_{n-1}(\cdots (g_1(\rz)). \cdots ))
\end{align}
The early layers of a generator represent high-level semantics such as the presence and layout of objects, while later layers encode lower-level pixel information such as edges and colors, as observed by Bau et al. ~\shortcite{bau2019gandissect}.  Therefore, to leave the semantic structure of $G$ unchanged, we divide it into a group of high-level layers $G_H$ containing layers $1$ through $h$ and fine-grained layers $G_F$ containing layers $h+1$ through $n$, so that $G(\rz) \equiv G_F(G_H(\rz))$.  This division is illustrated in \reffig{formula-adaptation}.  Only $G_F$ are adjusted when creating $G'$.  The early layers $G_H$ that decode the high-level structure of the image remain unchanged.  
In detail, we define $G_H$ and $G_F$ as:
\begin{align}
\rz_h \equiv \, G_H(\rz)  & \equiv g_{h}(g_{h-1}(\cdots g_{1}(\rz) \cdots )) \nonumber \\
G_F(\rz_h) & \equiv g_{n}(g_{n-1}(\cdots (g_{h+1}(\rz_h) \cdots ))
\label{eq:def-gfprime}
\end{align}
The choice of $h$ can be tuned experimentally; we have found it to be effective to choose $h = n - 5$ so that the fine-grained section contains a pyramid of four scales of convolutions.  To create a generator that faithfully reconstructs the target image $x$, we will update $G_F$.  However, directly updating the weights of fine-grained layers $G_F$ to match output to $\rx$ causes overfitting: when changed in this way, the generator becomes sensitive to small changes in $\rz$ and creates unrealistic artifacts.

Instead, we train a small network $R$ to produce small perturbations $\delta_i$ that multiply each layer's output in $G_F$ by  $1+\delta_i$. Each $\delta_i$ has the same number of channels and dimensions as the featuremap of $G_F$ at layer $i$.  This multiplicative change adjusts each featuremap activation to be faithful to the output image. (Similar results can be obtained by using additive $\delta_i$.)
Formally, we construct $G_F'$ as follows: 
\begin{align}
G_F'(\rz_h)  & \equiv g_{n}( (1+\delta_{n-1}) \odot g_{n-1}(\cdots ((1+\delta_{h+1}) \odot g_{h+1}(\rz_h) \cdots ))) \nonumber \\
G'(\rz)  & \equiv G_F'(G_H(\rz)).
\end{align}
The perturbation network $R$ learns to produce $\delta_i$ starting from a random initialization; $R$ takes no input.  \reffig{formula-adaptation} illustrates the architecture of this perturbation network. 

To further prevent overfitting, we add a regularization term to penalize large perturbations:
\begin{align}
\Ls_{\text{reg}} & \equiv \sum_{i=h+1}^{n-1} || \delta_i||^2
\end{align}.

\myparagraph{Overall optimization.}%
The overall optimization of $G'$ can now be summarized.  To learn $G'$, we fix the edited semantic representation $\rz_e$ and the unedited pixels $\rx \odot \mask_e$ that must be matched.  Furthermore, we fix all pre-trained layers of the generator $G$.

Our objective is to learn the parameters of perturbation network $R$  to minimize the following loss:
\begin{align}
\Ls = \Ls_{\text{match}} + \lambda_{\text{reg}} \Ls_{\text{reg}}.
\end{align}
We learn the network $R$ using the  standard Adam solver~\cite{kingma2014adam} with a learning rate of 0.1 for $1000$ steps.  The weight $\lambda_{\text{reg}}$   balances the magnitude of perturbations of $G'$ and closeness of fit to the target pixels. We empirically set it to 0.1.

The optimization takes less than 30 seconds on a single GPU.  Fewer steps can be used to trade off quality for speed.  While $1000$ steps achieve an average PSNR of $30.6$ on unedited reconstruction of a sample, it takes $100$ steps to achieve PSNR of $24.6$.

The computational form of adapting a network to a single image is inspired by previous work on deep image prior~\cite{ulyanov2018deep} and deep internal learning~\cite{shocher2018zero}, which have been shown to be effective at solving inpainting and super-resolution without any training data beyond a single image. However, our application is different; we initialize G with a generative model trained on a distribution of images in order to synthesize a semantic change encoded in the latent vector $\rz_e$, and then we train on a single target image in order to blend the semantic edits with the original image.

\subsection{Semantic Editing Operations: \texttt{GANPaint}} 
\lblsec{method-edit}
Our user interface (\reffig{ui_screenshot}) enables interactive editing of images that can be uploaded by users. Although computing $G'(\rz_e)$ requires some time, the interactive editor provides real-time previews $G'_w(\rz_e)$ using a $G'_w$ that is fast to apply.  The function $G'_w$ is set to a perturbation of the generator $G$ with weights that are optimized so that $G'_w(\rz) \approx \rx$.  Since $G'_w$ is derived only from the pixels of the unedited image $\rx$, it can be computed once when the image is uploaded and applied quickly for each edit.  Although $G'_w(\rz_e)$ introduces more visual artifacts compared to $G'(\rz_e)$, it gives a good indication of what the final rendering will look like at interactive rates. Currently, our web-based interface requires a server-side GPU for inference.

To demonstrate semantic editing, we use  Progressive GAN models~\cite{karras2018progressive} that can generate realistic scene images trained on the LSUN dataset~\cite{yu2015lsun}.  Once trained, this generator consists of 15 convolutional layers and produces images at a resolution of $256\times 256$ pixels. We closely follow methods of GAN Dissection~\cite{bau2019gandissect} to paint a photo by editing middle-level latent representations
located on the feature maps between the 4th and 5th layer. The representation $\rz \in \R^{ 8\times 8 \times 512}$ is a 512-channel tensor with featuremaps of size $8\times 8$. The GAN dissection toolkit~\cite{bau2019gandissect} can be used to identify a set of object types and concepts $C$ which we allow the user to insert, remove or change the appearances of objects in their images, as detailed below.

\begin{figure*}
\includegraphics[width=\textwidth]{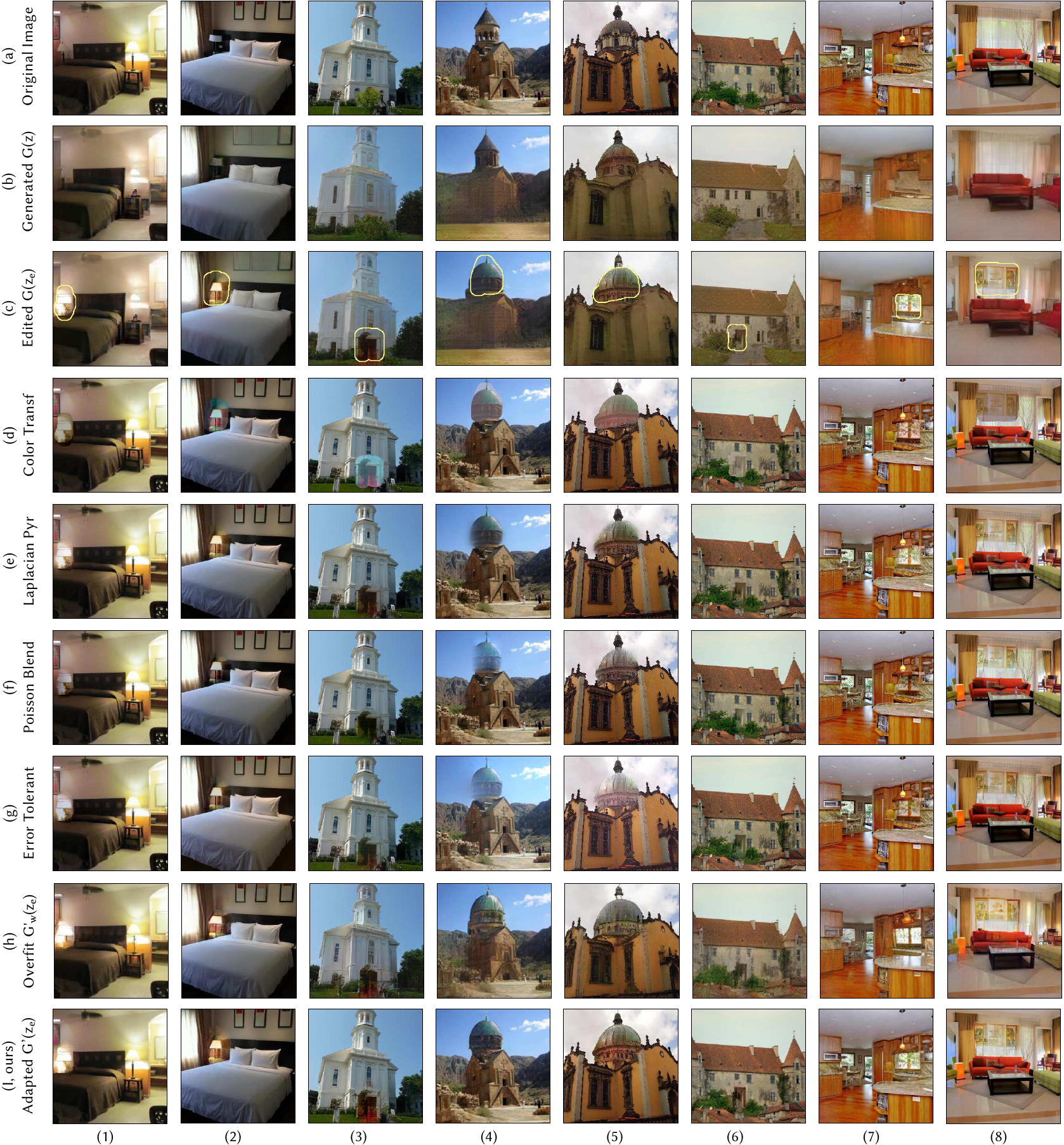}
\caption{Comparing our method to several compositing methods. From top to bottom: (a)~the original unmodified image (b)~its reconstruction by the unmodified generator; (c)~the generated image $G(\rz_e)$ after applying a semantic edit to the latent vector $z$; (d)~compositing the edited region from the edited generated (c) into the original image (a) using simple color transfer; (e)~Laplacian pyramid blending; (f)~Poisson blending; (g)~Error-tolerant image compositing; (h)~reconstruction of $z_e$ using a $G'_w$ where weights have been overfit to the unedited original image $x$; 
(i)~our method: reconstruction of $z_e$ using a $G'$ with activations adapted to match image $x$ outside the editing region.  Photos from the LSUN dataset~\citep{yu2015lsun}.}
\lblfig{composite-evaluation}
\end{figure*}

\myparagraph{Adding and removing objects.}
Using our \texttt{GANPaint} editor~\reffig{ui_screenshot}, users can select regions of an image they would like to edit.  To insert or remove an object belonging to class $c$ in a user-selected region $U \in \R^{8\times8}$, we form a channel mask $\alpha_c = (i_c \otimes U ) \in \R^{8\times8\times512}$, to select specific features in $z$ in the feature channels and locations that the user wishes to edit.
The feature channels $i_c \in \R^{512}$ relevant to class $c$ are obtained by analyzing the generator using~\cite{bau2019gandissect}, and $\otimes$ indicates a broadcasted outer product.

The edited representation $\rz_e = \edit(\rz)$ is computed as follows:
\begin{align}
\rz_{e} := \underbrace{(1 - \alpha_{c}) \odot \rz}_{\text{activations retained from } z} + \underbrace{\alpha_{c} \odot (s\, p_c)}_{\text{edited activations}}
\end{align}
In the above, the representation $\rz$ is blended with the feature vector $sp_c$, where $p_c \in \R^{8\times8\times512}$, is a spatially-expanded vector that represents the average activation of the object class $c$ over all images, constant in each $8\times 8$ channel, and $s$ is a scalar which controls how much to shrink or boost edited activations.  Setting $s = 0$ corresponds to removing class $c$ from the representation, whereas setting $s > 0$ corresponds to adding class $c$.

\myparagraph{Changing the appearance of objects.}
The same system can also alter the appearance of an object of class $c$. To do this, we simply use equation (9) with a value of $p_c$ derived from a \textit{\guidance} that has an appearance the user wishes to mimic. A simple way to accomplish this is to set the $i$-th component of $p_c$ to be the mean of the positive activations in the $i$-th channel of the latent features $\rz$ of the \guidance for channels related to class $c$; all other elements of $p_c$ are set to zero. See \reffig{style_variants} for an example of changing object appearances.
Taking means of positive activations is only one possible way to incorporate information about the \guidance's latent representation $\rz$ into $p_c$; we leave exploring other ways of copying attributes to future work.

\section{Experimental Results}
\lblsec{results}
We evaluate each step of our method.  In \refsec{compositing} we compare our image-specific adaptation method to several compositing methods, and in \refsec{ablation} we compare our method to two simplified ablations of our method.  In \refsec{qualitative}, we show qualitative results applying our image-specific adaptation $G'$ on both training set and in-the-wild test images. In \refsec{stylevariants} we demonstrate our method for changing the appearance of objects. In \refsec{inversion}, we evaluate our method for recovering the latent vector $\rz$ from an image $\rx$.

\subsection{Comparing Image-Specific Adaptation to Compositing}
\lblsec{compositing}

By adapting the generator to the original image, our method blends the generated edited image with the unedited parts of the photograph.  This scenario is similar to the task of image compositing, so we evaluate the performance of our method in comparison to several compositing approaches.  We compare our method to Color transfer~\cite{reinhard2001color}, Laplacian pyramid blending, Poisson editing~\citep{perez2003poisson}, and Error-Tolerant Image Compositing~\citep{tao2010error}.  In each case, we use the traditional compositing method to insert the edited image pixels generated by $G(z_e)$ from the edited region into the target photograph.

We asked 206 Amazon MTurk workers to compare the realism of the results of each compositing algorithm with our results.  For each method, 1200 pairwise quality comparisons were collected on 20 different edited images.  For each comparison, workers are shown the real photograph and the edited region as generated by $G(\rx_e)$, then asked which image from a pair is the most realistic insertion of the object into the photo. Top rows of Table~$\ref{tab:evaluation}$ summarize the results.  Qualitative comparisons are shown in~\reffig{composite-evaluation}(defg).

Workers find that our method yields more realistic results than traditional image blending approaches on average. For example, in the case \reffig{composite-evaluation}(c4), our method blends the new dome into the old building while maintaining edges, whereas traditional blending methods do not differentiate between boundary pixels that should be crisp edges from those that require smooth blending.
\begin{table}[h!]
\caption{AMT evaluation of compositing methods compared to our method: we report the percentage of users that prefer various other methods over ours.  Our method is also compared to the unadapted generator $G$ as well as a directly adapted generator $G'_w$ in which the weights have been fitted so $G'_w(z) \approx x$.}
\label{tab:evaluation}
\centering
\begin{tabular}{lc} 
\toprule
Method & {\small \% prefer vs ours} \\

\midrule
Color transfer~\cite{reinhard2001color}  & 16.8\% \\
Error-tolerant image compos.~\cite{tao2010error} & 43.6\% \\
Poisson blending~\cite{perez2003poisson} & 44.2\%  \\
Laplacian pyramid blending & 47.2\% \\
\bottomrule
Our method & 50.0\% \\
$G(z_e)$ without adaptation & 37.4\% \\
$G'_w(z_e)$, weights are fitted so $G'_w(z) \approx x$ & 33.1\% \\
\bottomrule
\end{tabular}
\end{table}

\subsection{Ablation Studies}
\lblsec{ablation}

To study the necessity of adapting $G'$ by perturbing activations, we compare our method $G'(\rz_e)$ to two simpler generator-based approaches.  First, we compare our method to $G(\rz_e)$ itself without any perturbations in G.  Second, we compare to an adapted $G'_w(\rz_e)$ in which the image is rendered by an adapted $G'_w$ with weights that have been optimized to fit the unedited image, so that $G'_w(\rz) \approx \rx$.  Although $G'_w$ is adapted to $\rx$, it is adapted to all pixels of the unedited image without regard to the editing region.  Results are summarized in bottom rows of Table~\ref{tab:evaluation}, and in \reffig{composite-evaluation}(ehi).  Note that the question asked for AMT raters shows the edited $G(\rz_e)$ as the example of the foreground edit that should be inserted in the photo.  Thus raters are primed to look for the editing changes that are demonstrated in image; this might result in ratings for $G(\rz_e)$ that could be higher than they would without this prompt.

Compared to the two ablations, our method produces results that are with fewer visible artifacts in the output image that are rated as much more realistic on average.  In particular, artifacts introduced by $G'_w$ are seen as high-frequency differences between the target image and the generated image, as well as additional color artifacts near the edited region.  These artifacts can be seen by zooming into the examples in \reffig{composite-evaluation}(h).

\subsection{Qualitative Results}
\lblsec{qualitative}

\begin{figure*}
\begin{center}
\includegraphics[width=1\textwidth]{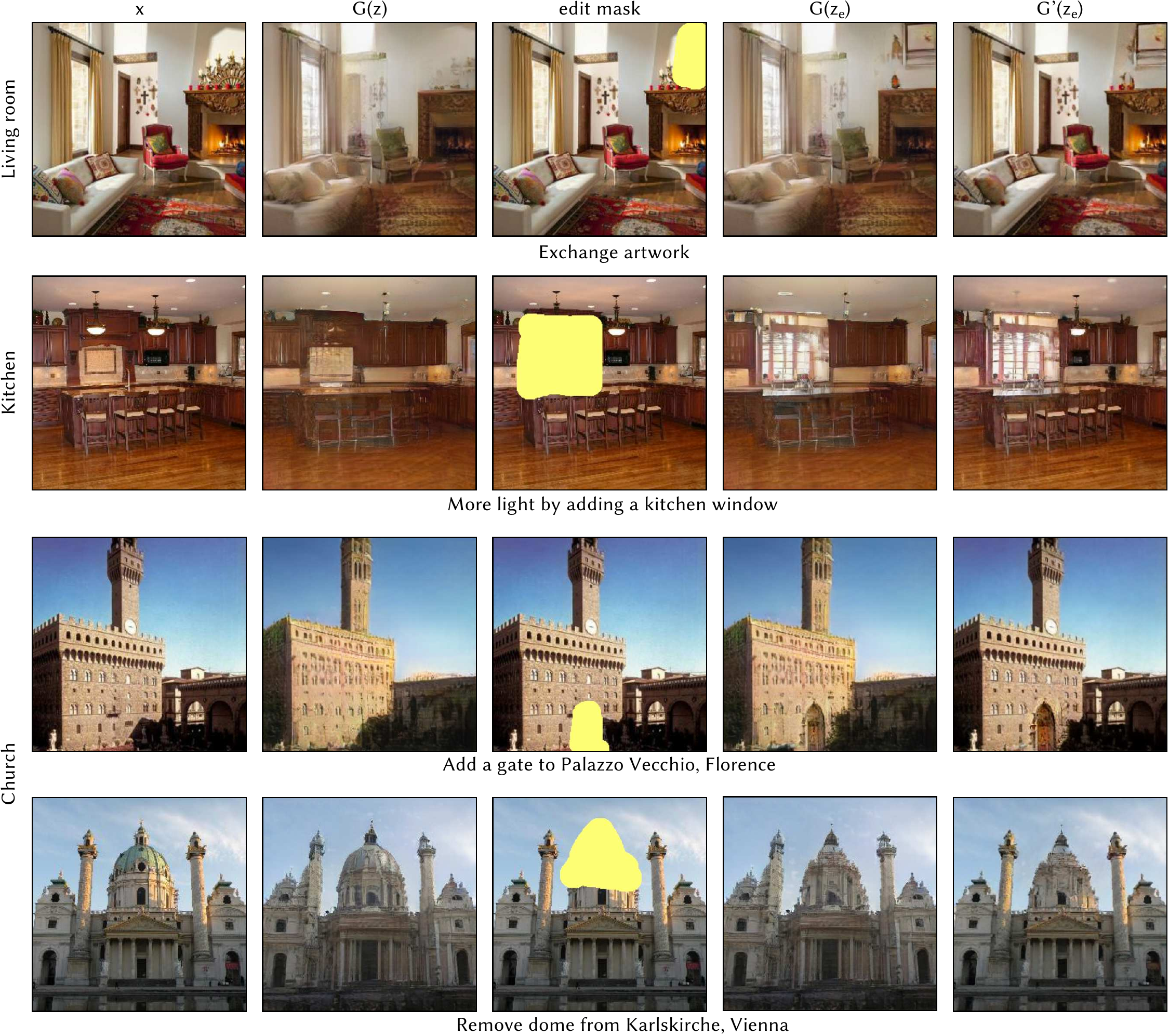}
\end{center}
\caption{Examples of editing work-flow. From left to right: input image $x$ is first converted to GAN image $G(z)$, edited by painting a mask, the effect of this mask edit can be previewed at interactive rates as $G(z_e)$. It can be finally rendered using image-specific adaption as $G'(z_e)$.  Photos from LSUN~\citep{yu2015lsun}.}
\lblfig{prominent_edits}
\end{figure*}

In this section, we show qualitative results of applying image-specific adaptation on natural photographs.

\myparagraph{Editing LSUN images.} In \reffig{prominent_edits}, we show several edits on images from the LSUN datasets~\cite{yu2015lsun}.  In the first column, we show the original unedited image, which is inverted (column two). The segmentation mask in column three indicates the edits requested by the user and the resulting GAN-generated image is shown in column four ($G(\rz_e)$).

To the right (column five) we show the results of applying image-specific adaptation on edits using the same user requests.  All examples are taken from the LSUN dataset on which the GANs were trained (living rooms, kitchens, outdoor church images).

\begin{figure*}
\includegraphics[width=\textwidth]{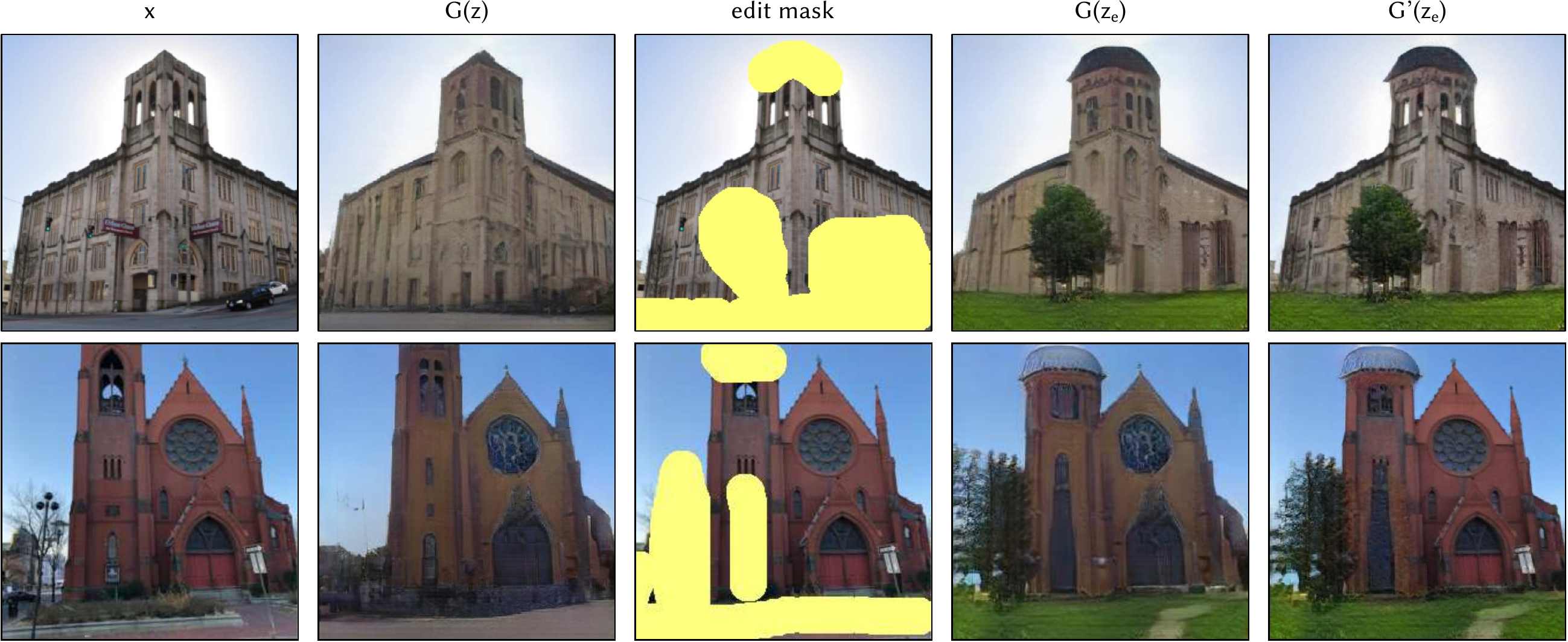}%
\vspace{-5pt}%
\caption{Applying our method on `in the wild' images.  These images are newly collected images that are not in the training set.  Images are edited using a GAN model trained on churches: new church images respond well to editing, even if they contain a few idiosyncratic objects (such as cars and lampposts) not modeled by the generator.  Photo of Urban Grace Church courtesy Visitor7 via \href{https://commons.wikimedia.org/wiki/File:Tacoma,_WA_-_Urban_Grace_Church_02.jpg}{Wikipedia} (cc-by-sa/3.0); photo of First Baptist Church by the authors.}
\lblfig{wild-church}
\end{figure*}

\myparagraph{Editing In-the-wild images.} In \reffig{wild-church}, we examine editing results on newly collected images that are not in the training set.  All edits are applied using a GAN trained on `outdoor church' images. %

As can be seen in the second column of \reffig{wild-church}, rendering of $G(z)$ reveals parts of images such as building shapes, doors, windows, and surfaces that are modeled by $G$. Many parts of the new church images are modeled, and our image-matching method is able to apply edit to these parts of the image in the presence of objects such as cars and lampposts that are not generated by $G(z)$.

\subsection{Style Variants}
\lblsec{stylevariants}
We demonstrate varying styles of inserted objects in \reffig{style_variants}. The variants are produced by using different source images ((a) and (b)) and by using the same image but different strengths of style adaptation (c). 
\begin{figure*}[t!h]
\begin{center}
\includegraphics[width=1\textwidth]{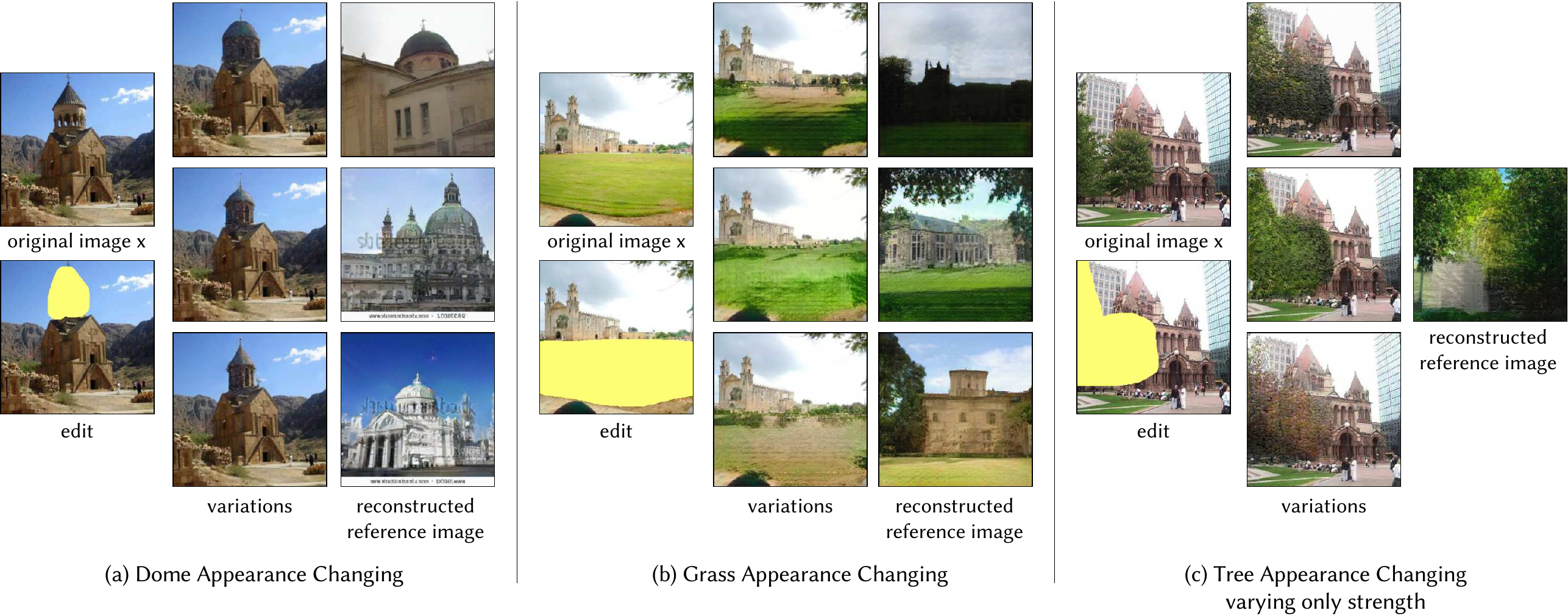}
\end{center}%
\vspace{-10pt}%
\caption{Changing the appearance of domes, grass, and trees. In each section, we show the original image $\rx$, the user's edit overlayed on $\rx$ and three variations under different selections of the \guidance. Additionally, we show reconstructions of the \guidance from G. In (c), we fix the \guidance and only vary the strength term $s$.  Photos from the LSUN dataset~\citep{yu2015lsun}.}
\vspace{-10pt}
\lblfig{style_variants}
\end{figure*}

\begin{figure}
\includegraphics[width=\columnwidth]{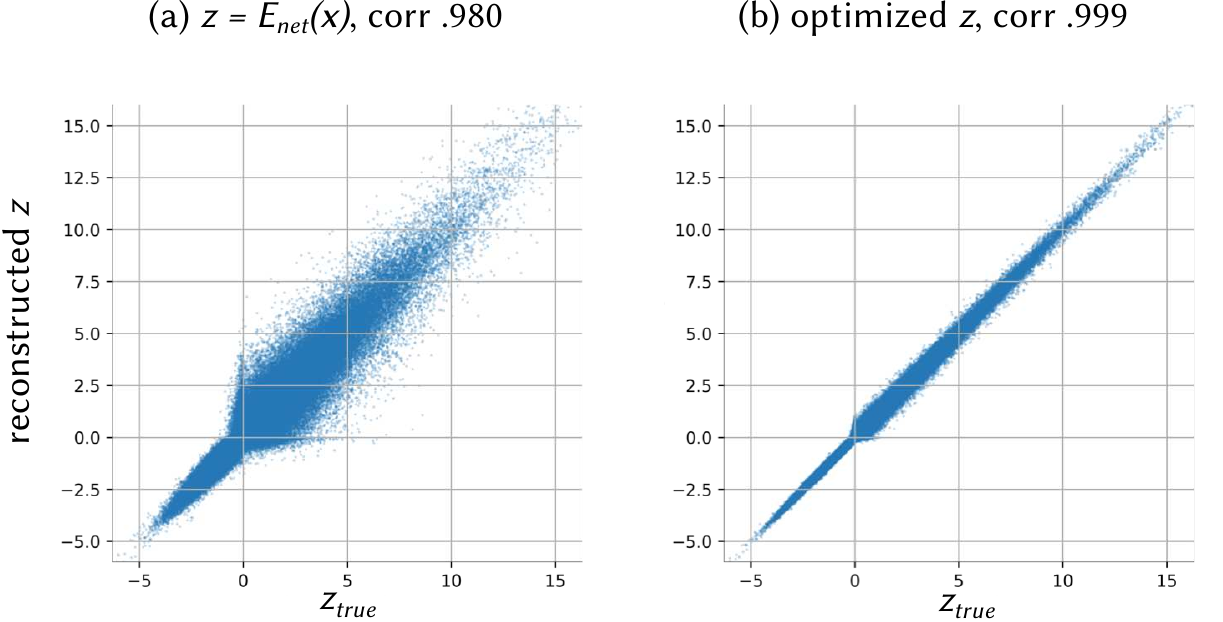}%
\vspace{-10pt}%
\caption{A generator $G$ can be inverted  accurately within its range.  Two inversion methods are tested on images generated by $\rx = G(\rz_{\true})$, and the components of the predicted $\rz = E(\rx)$ are plotted against the true components of the known $\rz$.  In (a), a Resnet-18 network is trained to calculate $G^{-1}$ and achieves good precision.  In (b), the network-computed results are used to initialize an optimization that further refines the calculation of $\rz$.  The result is a recovery of the true $\rz$ above a Pearson's correlation about 99.9\%.  Thus inversion works well for generated images.  However, this does not imply that these methods can invert real photos.  Images from the LSUN dataset~\citep{yu2015lsun}.}
\lblfig{inversion-scatter}
\end{figure}
\begin{figure}
\includegraphics[width=\columnwidth]{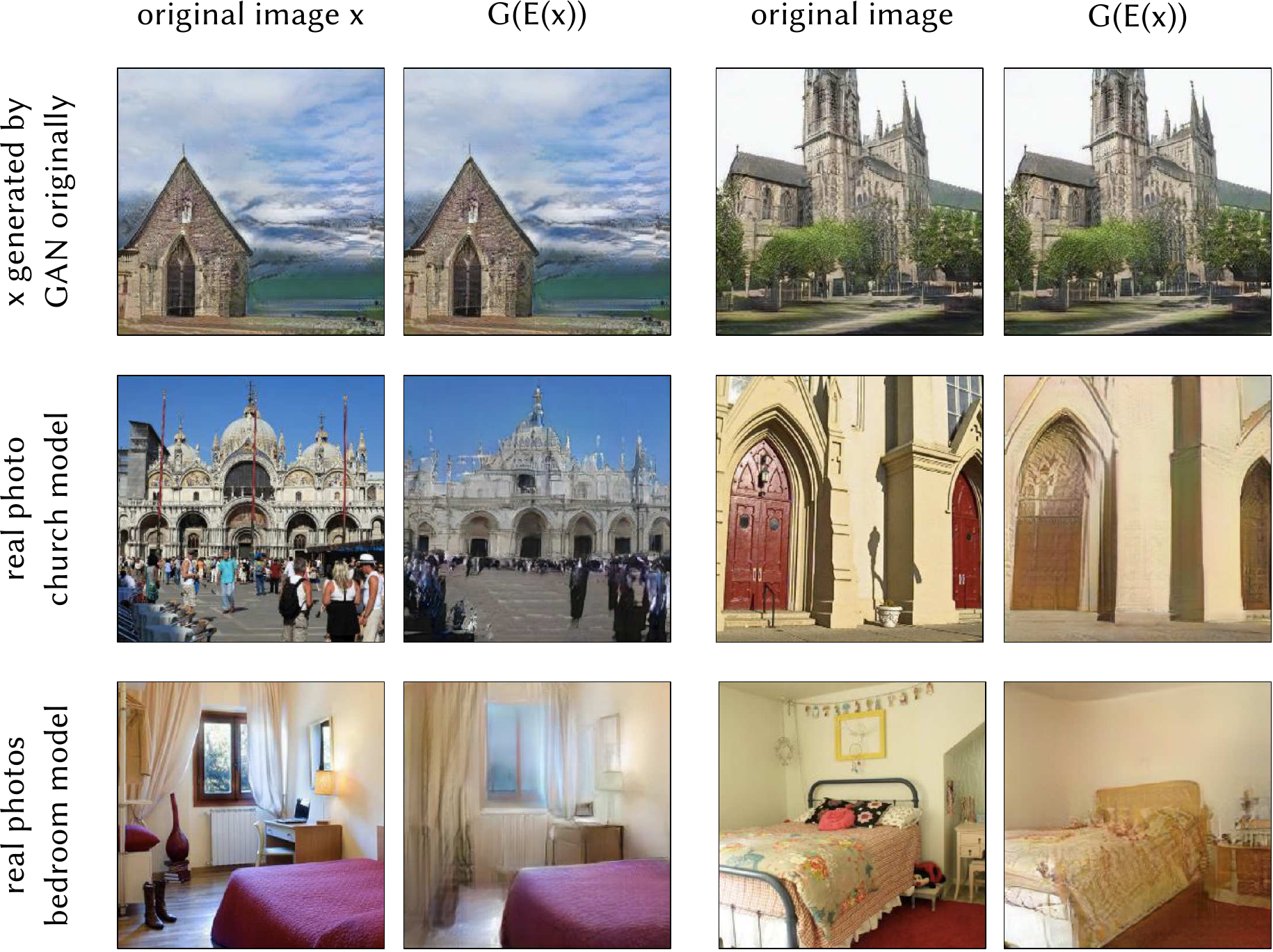}
\vspace{-10pt}
\caption{Recovery of latent vectors $\rz$ from images $\rx$: results can be compared qualitatively by rendering $G(\rz)$ and comparing to the original image $\rx$.  For images that were originally generated by the generator, our algorithm can recover $\rz$ that is quantitatively and qualitatively identical to its true value. However, this is not true for many real photos $\rx$.   Inverting and regenerating the image reveals details that the generator cannot represent; for the two models shown here, these details can include people and objects that are not usually seen in the data set.  Photos from LSUN~\citep{yu2015lsun}.}
\vspace{-5pt}
\lblfig{inversion-qualitative}
\end{figure}
\subsection{Recovering the Latent Vector z}
\lblsec{inversion}
Our method depends on starting with a latent vector $z$ that is able to approximate the user's photo.  Here we report our results in recovering an inverse $E\approx G^{-1}$. We find $z$ to optimize the latent vector $\rx \approx G(\rz)$ in the following two steps.

First, we train a network $\Enet(\rx)$ to minimize the mean loss \refeq{opt} when inferring $\rz$ from $\rx$.  We find that ResNet-18 is effective at computing a good estimate for $\rz$.  We can quantitatively evaluate the accuracy of $\Enet$ at recovering the correct $\rz$ by testing it on images $\rx = G(\rz_{\true})$ for which the true latent vector $\rz_{\true}$ is known.  The results are shown in \reffig{inversion-scatter} (a).

Next, we apply gradient descent to optimize the latent vector $\rz$ using the same objective function (\refeq{opt}).  Our case, $\rz = \rz_4$ is a fourth-layer representation of an underlying progressive GAN, which means that we do not have explicit statistics for regularizing this representation.  We have found that an effective implicit regularizer for $\rz_4$ can be obtained by optimizing over the earlier layers $\rz_1, \rz_2, and~\rz_3$ rather than $\rz_4$ itself.  Using this approach results in the recovery of nearly perfect $\rz_4$, as shown in \reffig{inversion-scatter} (b).
We use $E(\rx)$ to denote the result of both steps of the optimization. \reffig{inversion-qualitative} shows a qualitative comparison of reconstructions $G(E(\rx))$ with input images $\rx$ for two pairs of images.  We observe that, for $\rx$ that are generated by $G$, the inversion is nearly perfect: the reconstructed image is indistinguishable from the original. \reffig{inversion-qualitative} also shows reconstructions $G(E(\rx))$ for real images $\rx$ that are \textit{not} generated by $G$. In this setting, the inversions are far from perfect. Furthermore, the failure of $E$ on other images reveals the types of images that $G$ cannot reconstruct, and provide insight into the limitations of the generator.  For example, in a model trained on outdoor scenes of churches, the inversions cleanly drop out people and vehicles, suggesting that this model cannot represent these types of objects well. The model trained on bedrooms drops out certain types of decorations.

\begin{figure}
\includegraphics[width=0.47\textwidth]{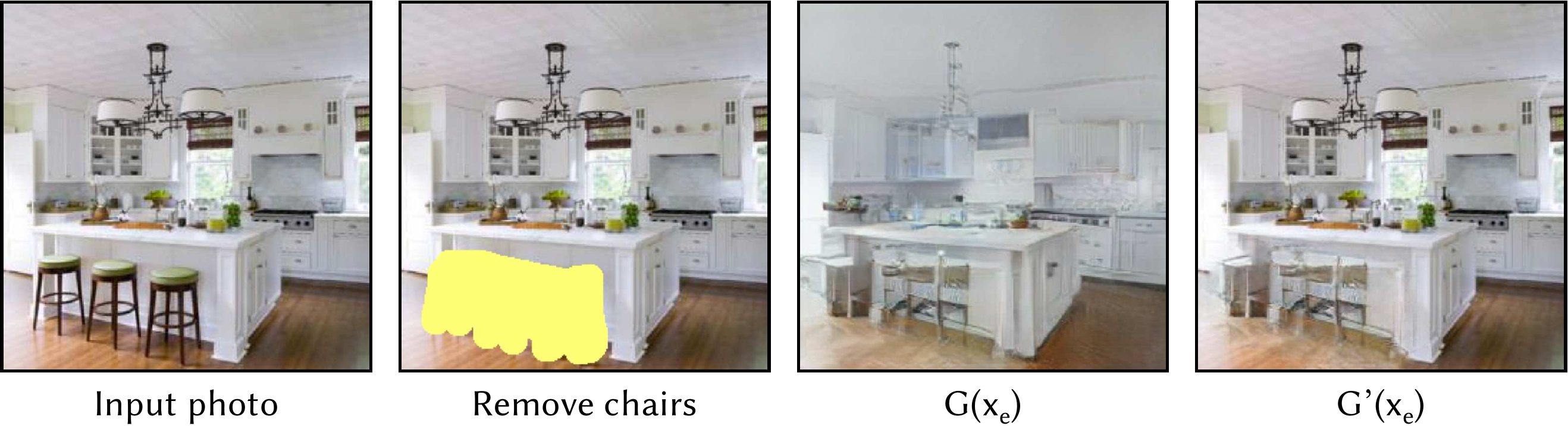}
\caption{\emph{Failure cases.} Some latent vector space operations are difficult to disentangle from undesired effects. For example, in this sequence, a user has attempted to use a latent space manipulation to remove all the chairs from the image.  In the generated version of the image, it can be seen that when the chairs are removed, visual remnants remain mixed with the space.  Disentanglement of latent vector directions remains a core challenge for this family of methods.  Photo from the LSUN dataset~\citep{yu2015lsun}.}
\lblfig{entangled-chairs}
\vspace{-10pt}
\end{figure}

\section{Limitations and Discussion}
\lblsec{discussion}
Although our method for adapting a generative network allows various semantic photo manipulations, many challenges remain. First, our method requires an optimization be run after each edit, which takes about 30 seconds on a modern GPU. This time can be reduced by applying fewer optimization steps and incrementally optimizing during editing, or by using a full-image adaptation $G'_w$ of the generator that provides a adaptation of the generator that does not depend on the editing region.  Although these faster methods introduce more artifacts than our full method, they can provide useful preview of the results at interactive speeds.

Second, another challenge is that latent spaces learned by deep neural networks are not fully disentangled, so finding vector operations to express a user intent can remain challenging. Adding or removing an object from a scene may have some interactions with unrelated objects. Some examples of undesirable interaction are illustrated in \reffig{entangled-chairs}; in these examples, chairs are reduced by zeroing chair-correlated components in the representation, but the rendered results do not cleanly remove all parts of the chairs, and distorted parts that resemble chair legs remain in the result. We also note that our method for varying the appearances of objects can be brittle; certain classes of objects (such as trees) vary more than other classes (such as domes or skies). For example, in Figure (10) we found that dome shape, but not color, could be varied.

Lastly, the quality and resolution of our current results are still limited by the deep generator that we are using~\cite{karras2018progressive}. For example, in \reffig{wild-church}, the GAN-synthesized images omit many details such as cars and signage; our method cannot add such objects to an image if they are not modeled by the generator. But our method is not coupled with a specific generative model and will improve with the advancement of models with better quality and resolution (e.g., concurrent work~\cite{karras2019style,brock2019large}). %

Nevertheless, we have found that in many computer graphics applications the learned natural image prior can help produce more realistic results with less user input. Our work presents a small step towards leveraging the knowledge learned by a deep generative model for semantic photo manipulation tasks.

\begin{acks}
We are grateful for the support of the MIT-IBM Watson AI Lab, the DARPA XAI program FA8750-18-C000, NSF 1524817 on Advancing Visual Recognition with Feature Visualizations, and a hardware donation from NVIDIA.
\end{acks}
\clearpage

\bibliographystyle{ACM-Reference-Format}

\bibliography{main}

\end{document}